**Deep-Learning-based Fast and Accurate 3D CT Deformable Image Registration in Lung Cancer**

Running title: Deep-learning based DIR

Yuzhen Ding, PhD[1], Hongying Feng, PhD[1], Yunze Yang, PhD[1], Jason Holmes, PhD[1], Zhengliang Liu, MS[2], David Liu[3], William W. Wong, MD[1], Nathan Y. Yu, MD[1], Terence T. Sio, MD, MS[1], Steven E. Schild, MD[1], Baoxin Li, PhD[4], Wei Liu, PhD[1]

[1]Department of Radiation Oncology, Mayo Clinic, Phoenix, AZ 85054, USA

[2]Department of Computer Science, University of Georgia, Athens, GA 30602, USA

[3]Athens Academy, Athens, GA 30602, USA

[4]School of Computing and Augmented Intelligence, Arizona State University, Tempe, Arizona, USA 85281

Corresponding author: Wei Liu, PhD, Professor of Radiation Oncology, Department of Radiation Oncology, Mayo Clinic Arizona, 5777 E. Mayo Boulevard, Phoenix, AZ 85054; e-mail: Liu.Wei@mayo.edu.

**Acknowledgments**

This research was supported by the National Cancer Institute (NCI) Career Developmental Award K25CA168984, Arizona Biomedical Research Commission Investigator Award, the Lawrence W. and Marilyn W. Matteson Fund for Cancer Research, and the Kemper Marley Foundation.

## Conflicts of Interest Notification

Terence T. Sio provides strategic and scientific recommendations as a member of the Advisory Board and speaker for Novocure, Inc., Catalyst Pharmaceuticals, Inc. and Galera Pharmaceutics, which are not in any way associated with the content presented in this manuscript.

## Ethical considerations

This research was approved by the Mayo Clinic Arizona institutional review board (IRB, 13-005709). The informed consent was waived by IRB protocol. Only CT image and dose-volume data were used in this study. All patient-related health information was removed prior to the analysis and also publication of the study.

**Abstract**

**Background:** Deformable Image Registration (DIR) is an essential technique required in many applications of radiation oncology. However, conventional DIR approaches typically take several minutes to register one pair of 3D CT images and the resulting deformable vector fields (DVFs) are only specific to the pair of images used, making it less appealing for clinical application.

**Purpose:** In some proton therapy facilities, patient alignment relies on two 2D orthogonal kV images, taken at fixed, oblique angles, as no 3D on-the-bed imaging is available. The visibility of the tumor in kV images is limited since the patient's 3D anatomy is projected onto a 2D plane, especially when the tumor is behind high-density structures such as bones. This can lead to large patient setup errors. A solution is to reconstruct the 3D CT image from the kV images obtained at the treatment isocenter in the treatment position.

**Methods:** An asymmetric autoencoder-like network built with vision-transformer blocks was developed. The data was collected from 1 head and neck patient: 2 orthogonal kV images (1024x1024 voxels), 1 3D CT with padding (512x512x512) acquired from the in-room CT-on-rails before kVs were taken and 2 digitally-reconstructed-radiograph (DRR) images (512x512) based on the CT. We resampled kV images every 8 voxels and DRR and CT every 4 voxels, thus formed a dataset consisting of 262,144 samples, in which the images have a dimension of 128 for each direction. In training, both kV and DRR images were utilized, and the encoder was encouraged to learn the jointed feature map from both kV and DRR images. In testing, only independent kV images were used. The full-size synthetic CT (sCT) was achieved by concatenating the sCTs generated by the model according to their spatial information. The image quality of the synthetic CT (sCT) was evaluated using mean absolute error (MAE) and per-voxel-absolute-CT-number-difference volume histogram (CDVH).

**Results:** The model achieved a speed of 2.1s and a MAE of <40HU. The CDVH showed that <5% of the voxels had a per-voxel-absolute-CT-number-difference larger than 185 HU.

**Conclusion:** A patient-specific vision-transformer-based network was developed and shown to be accurate and efficient to reconstruct 3D CT images from kV images.

**Keywords:** deformable image registration, 3D lung CT images, deep neural networks

## 1. Introduction

Image registration aims to find the spatial relationship between two or multiple sets of images and is usually formalized as the optimization of a function balancing the similarity between images (either in intensity, topology, or both)[1]. Compared to rigid image registration (RIR), deformable image registration (DIR) attempts to find the voxel-specific spatial relationship between two or multiple sets of images. Therefore, DIR has far more flexibilities than RIR and can be used in more complicated clinical scenarios such as images with large anatomical structure changes. DIR has been extensively used in radiation therapy[1] such as automatic segmentation[2,3], mathematical modeling[4-7], functional imaging[8-10], and dose deformation[11-16].

Over the years, many conventional DIR approaches have been developed and adopted clinically. The conventional DIR approaches can be broadly categorized into two categories: parametric[6,7,17] and non-parametric models[18-21]. The parametric model generates DVFs as a linear combination of its basic functions. The B-spline model[22-25] is an example of such parametric models and it can handle the local change of a voxel by linear regression from nearby voxels within a certain distance. This property significantly reduces the computation time and memory required. For example, Shekhar et al.[26] proposed a DIR framework for auto-segmentation. The framework consists of a B-spline-based transformation model, mean squared difference-based image similarity measure, and a downhill simplex algorithm as the optimization scheme. It achieved fewer than 120HU and 135HU mean squared difference for lung and abdomen patients, respectively. Yet, the results can only be used for CTs with either breath-holding or respiratory gating, which limit its wide applications in clinics. In contrast, non-parametric models such as demons-based[18-21] methods calculate transformation vectors of all voxels, thus achieving more accurate DVFs, but requiring more computation time and memory than the parametric models. For example, Reed et al[27] achieved an average of 1.3 mm mean displacement in auto-segmentation for 10 patients using an accelerated ‘‘demons’’ algorithm,[28] which adds a HU number gradient similarity term and a transformation error term into the demons' energy function, and uses the limited Broyden-Fletcher-Goldfarb-Shanno (L-BFGS) algorithm[29] to automatically determine the iteration number, thus accelerating the algorithm. However, it also requires the patients to have a similar body mass index (BMI)[29], which also potentially limits its application clinically.

Modern radiation therapy is increasingly sophisticated with more beam delivery techniques such as intensity modulation and/or volumetric modulation, including intensity modulated X-ray-based radiation therapy (IMRT)[30-33], volumetric modulated arc therapy (VMAT)[34], and intensity-modulated proton therapy (IMPT)[35-43]. IMPT enjoys distinct advantages in terms of high conformality of target coverage and superior organs-at-risk (OARs) protection owing to its high flexibility at the beamlet level in treatment planning and dose delivery[35-38]. However, it is also extremely sensitive to proton beam range, patient setup uncertainties, intra- and inter- fractional anatomical changes.[38,44-78] The concept of adaptive radiotherapy (ART)[14,79,80] has been introduced to account for anatomical changes during treatment course. For ART, patients under treatment require periodic verification imaging during treatment course to obtain information about their internal anatomical changes. However, the potential gain of ART is at the cost of increasing clinical workload, such as CT deformation, contour propagation, and dose deformation. Those clinical tasks all depend on the availability and quality of DIR. Unfortunately, it typically takes minutes for the conventional DIR approaches to register one pair of 3D CTs and the resulted deformable vector fields (DVFs) are not generalized to other CT images, even when they are similar or from the same patient, hence greatly limiting its further applications in ART, which is very time sensitive. Moreover, the frequency of re-planning is significantly higher in IMPT than IMRT/VMAT (for example, for head and neck cancer, 20-25% for IMRT/VMAT and 45-50% for IMPT). This makes the same tasks even more labor intensive in proton clinics. Therefore, the undesired patient breaks allowing for tumor cell repopulation might take place at busy clinics due to insufficient resources.

Recently, several deep learning-based methods have been developed to speed up DIR in medical image analysis[81-83]. Yang et al.[81] proposed a two-steps deep learning framework for predicting the momentum parameterization for the large deformation diffeomorphic metric mapping (LDDMM) model. The proposed deep learning framework consists of two auto-encoder networks with the same architecture, in which the first auto-encoder is used to estimate the initial patch-wise momentum and the second one further tunes the initial patch-wise momentum. Although the proposed method is much faster comparing to the conventional DIR approaches, the computational complexity is higher than a typical single-step deep learning network. Besides, since it has two cascade networks, the symmetrical error may accumulate as the layers go deeper. Balakrishnan et al.[82] proposed a UNet-like model termed as VoxelMorph to learn the DVFs from pairs of magnetic resonance images (MRIs) (i.e., moving images and fixed images), then the generated DVFs and moving images go through a non-learnable spatial transformation to form the final generated warped images that resemble the fixed images. The VoxelMorph can achieve comparable performance as the state-of-the-art conventional DIR methods, whereas it is orders of magnitude faster. Thus, it has been widely used in medical image analysis. Most of these methods have been proposed for MRIs, which typically have high-resolution and rich anatomical information, whereas in radiation therapy the commonly used image modality is CT with a relatively low resolution. Vos et al.[83] proposed a deep learning image registration (DLIR) framework for unsupervised affine and deformable image registration. It uses convolutional layers to predict the B-spline control points in each of the three directions, then the DVFs are generated from the estimated control points by B-spline interpolation, which is implemented by transpose convolutions. Although the DLIR can be applied to both MRIs and CTs, when it is used for CT images, it requires a large training dataset to train the model. Moreover, the trained model can only be used for 4D CTs with only intra-fractional anatomical changes considered, which limits its clinical use. Another deep learning-based DIR approach was proposed by Zhao et al.[84] by cascading multiple Volume Tweening Network (VTN) networks to recursively generate coarse-to-fine DVFs. Typically, the more the cascades are, the more accurate the generated DVFs are. However, the number of the cascades is bounded by the GPU memory, and a large amount of data is required to train such a large-scale network, which is challenging for tasks involving medical images.

To address the aforementioned challenges for the deep learning-based DIR approaches to be used in CTs (e.g., dependence of large training dataset, limitation of 4D CTs, and requirements of high resolution, which is not available in CTs), we proposed several additional loss terms in the objective function of VoxelMorph as well as a random masking strategy to greatly improved the quality of the synthetic CT images (a similar idea has been adopted by He et al.[93] to significantly accelerate training speed as well as improve the classification accuracy.), yielding an efficient, accurate, and generalizable deep-learning based DIR method for CTs.

The contributions can be summarized as follows:

1. We proposed a novel VoxelMorph-based framework for inter-fractional lung DIR. Different from conventional DIR approaches that are only specific to the images used, our framework can be generalized to any images of any independent patients once the model is trained. Thus, it is more practical and versatile.

2. A new random masking strategy was proposed to significantly reduce artifacts in the deformed images due to intrinsic low resolution of the CT images compared with MRIs. In addition, we investigated the functionalities of different loss terms used in the model training and used weighted mean absolute error (wMAE) and structural similarity index matrix (SSIM) loss (optional) to bridge the gap between CT images and MRIs, the latter has been well studied in deep learning-based DIR. Thus, the image quality of the deformed CTs is further improved.
3. Dedicated pre- and post-process methods are proposed to standardize all the CT images used in this work. Then, as a demonstration, we constructed a novel diversified inter-fractional lung CT dataset consisting of approximate 200 pairs of such standardized CT images collected from patients treated by both proton therapy and photon therapy. Such a dataset can be used to evaluate the performance of not only the DIR approaches, but also other related tasks. In the meantime, the proposed pre- and post-process methods can be applied to other medical images (e.g., head and neck CT images, MRI, etc.).
4. Our methods achieved the state-of-the-art performance in terms of time efficiency, high reconstructed image quality, indistinguishable dose distribution difference calculated between the ground-truth and deformed CTs, and good Gamma passing rates.

## 2. Methods

To address the drawbacks of the conventional DIR approaches, such as low accuracy and large time consumption, we propose to train a deep-learning-based model for the deformable vector fields (DVFs) with VoxelMorph, which is a general-purpose library for deep-learning-based tools for registration and deformations and includes two additional loss terms that focus on voxel-level similarity and structure-level similarity, respectively. We also introduce a new training strategy that can alleviate the artifacts in low resolution images (i.e., CT images) thus achieving accurately warped CT images. In section 2.1, we describe the data collection and data preprocessing. Section 2.2 introduces the overall structure of the proposed method and in section 2.3, the training process and validation process of the proposed method are elaborated. Statistical analysis is included in section 2.4.

## 2.1. Data Pre-processing

The initial CT ($iCT_o$) and verification CT ($vCT_o$) of 114 lung cancer patients treated at our institution were retrospectively selected, among which the CT images from 104 patients were used for training and 10 were used for testing. Each patient had one initial CT and 1-4 verification CTs, forming a training dataset of 192 pairs of CT images and a testing dataset of 10 pairs of CT images. In the training dataset, 101 pairs were collected from 67 patients treated with photon therapy whereas the other 91 pairs were collected from 37 patients treated with proton therapy. Among the 10 testing patients, 7 patients were treated with photon therapy, while the other three were treated with proton therapy.

As the collected CT images were captured at various times and by various CT simulators, the CT images may be different due to anatomical changes and various configurations of the different CT simulators. To make sure that the dataset was consistent, data preprocessing was conducted as follows. First, we used the iterative metal artifact reduction (iMAR) algorithm[85], which is integrated in the commercial software for the CT simulator, to remove artifacts caused by metal implants. Then, rigid registration and center-cropping were applied to all the CT images using the following technique: we first randomly picked one CT image, where the regions of interest (ROIs) were roughly located in the center of the 3D CTs. We regarded this CT image set as the reference CT (*rCT*). Then, we registered (rigid) all other CTs to *rCT* using the Insight Tookit (ITK)[86] such that all CTs had the same resolution of $2 \times 1.26 \times 1.26$ mm$^3$, the same dimension size, etc. Next, we center-cropped all CTs to a dimension size of $136 \times 384 \times 384$ to exclude the non-informative areas from this study as well as to alleviate the memory burden in training. We manually selected a fixed center-cropping region instead of using the BODY contour since the BODY contour varied from patient to patient, and in some proton plans the BODY contour contained the digital couch. Last, we normalized all the CT numbers to values approximately around 0 to 1 by using a uniform shift of 1,000 and a fixed denominator of 3,000. The preprocessed initial CT (*iCT)* and verification CT (*vCT*) were then used for the model training and validation. The workflow of both the data pre-processing and data post-processing steps is shown in Figure 1.

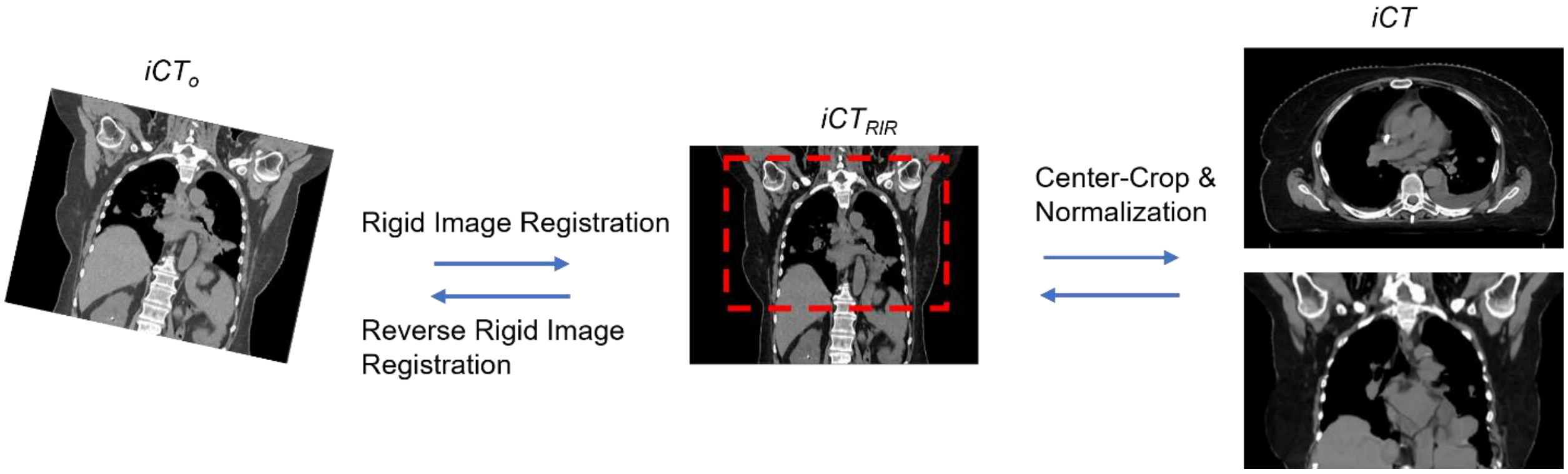

Figure 1. Illustration of the data pre-processing and post-processing steps. $iCT_o$ represented the raw data, $iCT_{RIR}$ represented the CT images after RIR was applied to $iCT_o$ and $iCT$ were the images that we used in the model training and validation.

### 2.2. Overview of the proposed framework

An overview of the proposed framework is illustrated in Figure 2 and the detailed model architecture is shown in Figure 3. The input for the model is a pair of CT images, *iCT* and *vCT*, that are taken at different time points (usually several weeks apart). The backbone of the model is VoxelMorph, which is a UNet-like deep neural network architecture and the output from the network is the DVFs. Finally, the *vCT* undergoes a spatial transformation based on the derived DVFs to form the final output – the *sCT* (Fig. 2).

Since the 3D lung CT images have different resolutions as well as different dimension sizes as the MRIs used in the original VoxelMorph model, the kernel size, stride, and other parameters are changed accordingly to make sure that the CT images and the network are compatible. To be more specific (Fig. 3), 3D convolutional layers are used with a kernel size of 3 and a stride of 1 in both the encoder and decoder. A LeakyReLU[87] layer with a parameter of 0.2 was applied right after each convolutional layer. The convolutional layers together with down-sampling across different layers allow us to capture the hierarchical features, which are derived from the input CT image pairs. Similarly, the decoder learns the DVFs from both the hierarchical features extracted by each layer in the encoder and the previous layer in the decoder. To deal with the odd number of the feature maps in the deepest layer of the encoder and decoder, we randomly duplicated one of the feature maps and concatenated it with the original feature maps (the second layer in the decoder), thus the number of the feature maps is consistent in both the encoder and decoder. Finally, the output from the network, i.e., the DVFs, were applied to the *vCT* image to generate the *sCT* image through spatial transform, in which the voxel location is first calculated then followed by linear interpolation. The *sCT* quality is evaluated with *iCT* as the ground-truth.

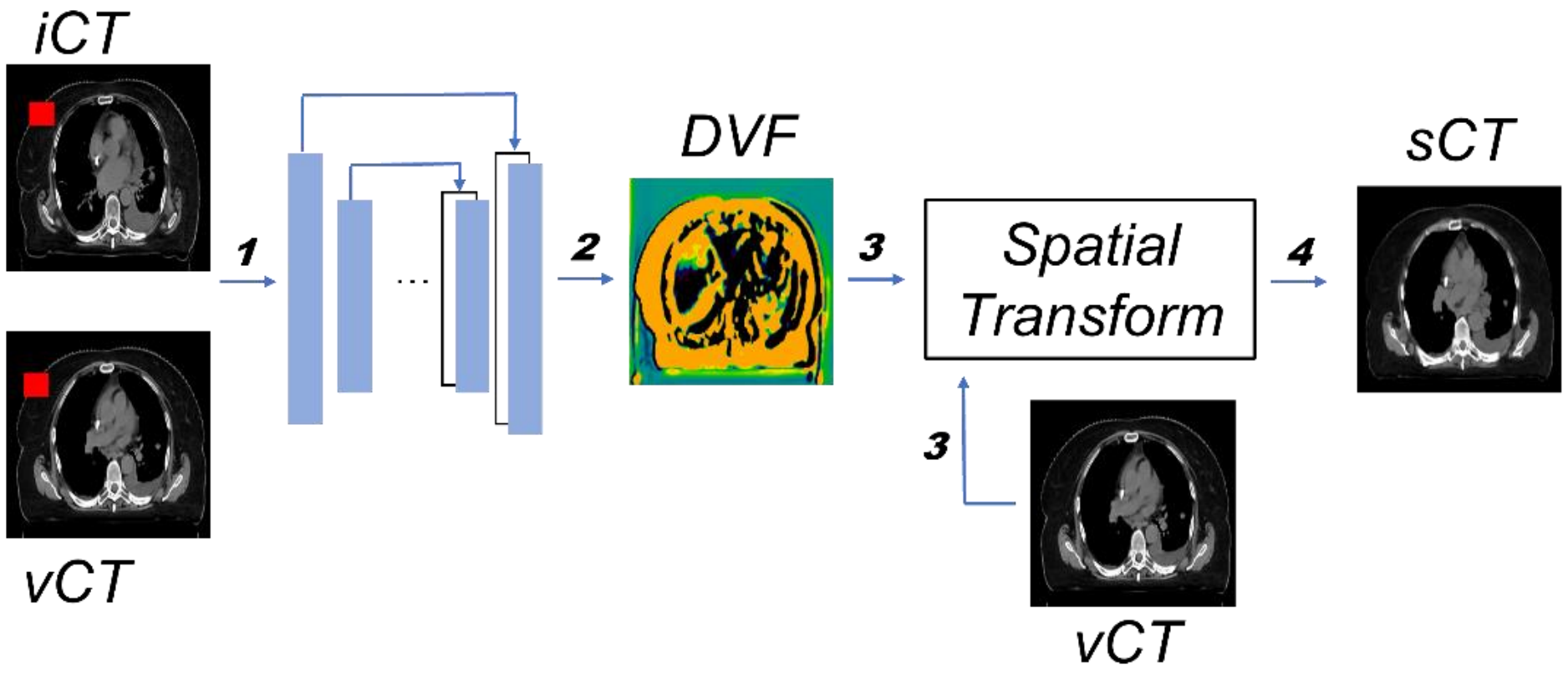

Figure 2. Overview of the proposed workflow. The inputs consist of both *iCT* and *vCT* (step 1). Both *vCT* and DVF, which is generated by the model (step 2), will go through the spatial transformation (step 3) to obtain the final output *sCT* (step 4). The training and inference path have been indicated by bolded numbers. Note that the random mask (red rectangle block in both *iCT* and *vCT*) is only applied in the training stage. More details about random mask will be introduced in section 2.3.

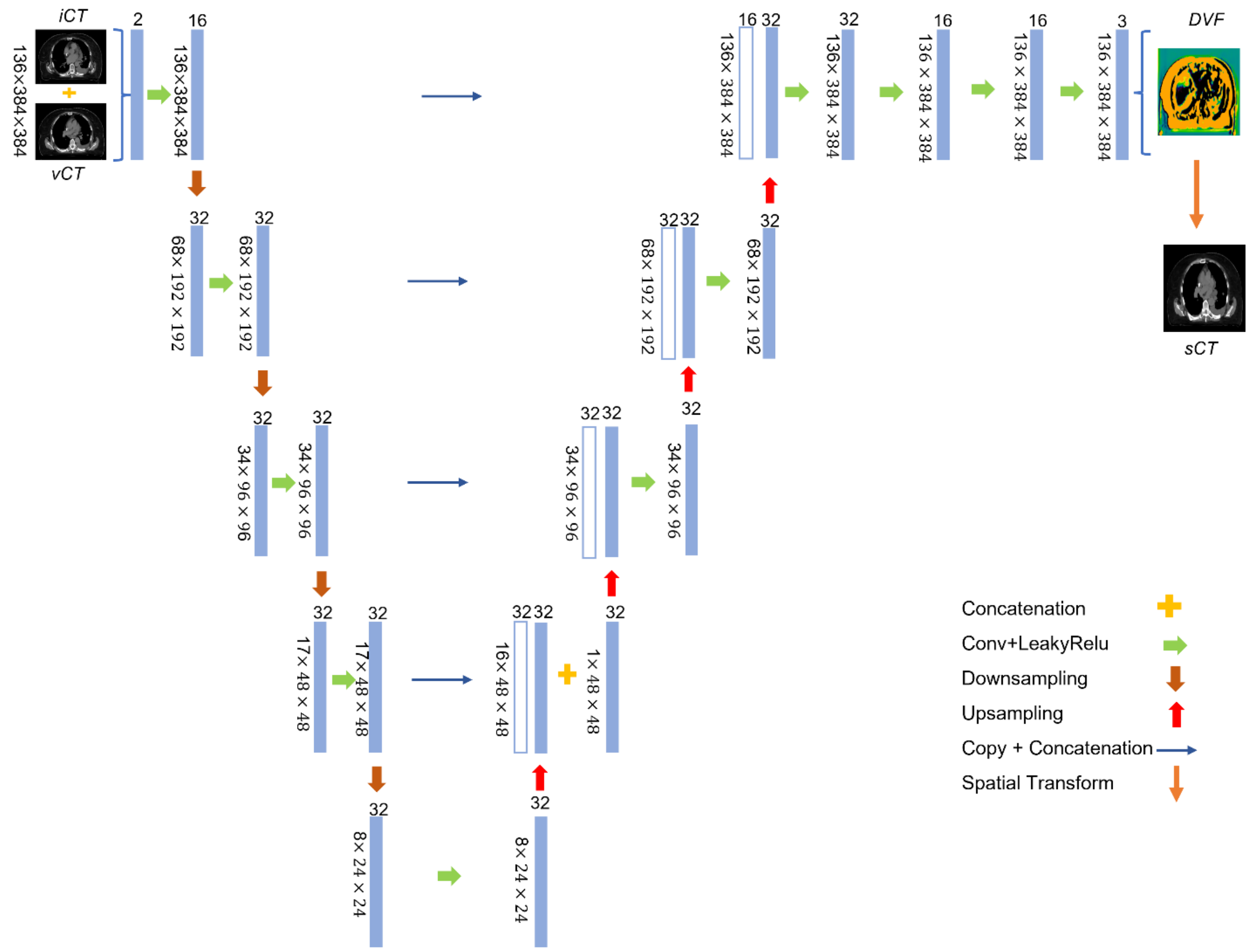

Figure 3. Details of the neural network architecture. Every block represents one layer, the value on the left is the input image size, whereas the value on the top indicates the number of feature maps.

## 2.3. Training and validation protocols of the proposed framework

As shown in Figure 2 and Figure 3, the base architecture is the VoxelMorph, which is a UNet-like structure that was proposed for DIR of the MRI images in head and neck. Although the vanilla VoxelMorph works well for DIR of the MRI images in head and neck, its performance greatly degenerated when directly applied to lung CT images. A few reasons contribute to such a degeneration: 1) The MRI images typically have a much higher resolution than the CT images and thus the former will let the model capture more voxel-wise details; 2) The number of the MRI images used in the previous study are much larger than the number of the CT images used in this study to which the model can easily overfit; 3) The lung disease site has larger variation across different patients and within one patient due to inter/intra-fractional anatomy changes than the head and neck disease site. Therefore, to address the above-mentioned challenges, we proposed a new training strategy and a new loss function.

In the training stage, we proposed to use the weighted per-voxel HU number mean absolute error (MAE) (*wMAE*) loss to measure the voxel-wise similarity between the ground-truth *iCT*s and synthetic *sCT*s. The definition of the *wMAE* is:

$$L_{wMAE}(iCT, sCT) = w_{p_{HU}} \sum_{p\in\Omega} |iCT(p) - sCT(p)| \quad (1)$$

where $w_{p_{HU}}$ represents the HU number of the voxel $p \in \Omega$. Different from a plain MAE loss, the weight of the similarity loss of each voxel is proportional to the corresponding HU number of the voxel. Hence, the voxels in structures with higher HU number, e.g., bone, were assigned with larger weights than other voxels. This, together with other loss terms help diminish the appearance of high HU number artifacts.

Following VoxelMorph, we also applied the smooth loss term to the generated DVFs, making it physically realistic. The smooth loss term is defined in Equation (2) as follows:

$$L_{smooth}(DVF) = \sum_{p\in\Omega} \|\nabla g(p)\|^2 \quad (2)$$

where $\nabla g(p) = \left(\frac{\partial g(p)}{x}, \quad \frac{\partial g(p)}{y}, \quad \frac{\partial g(p)}{z}\right)$ is the spatial gradients $\nabla g$ of the voxel $p$. To simplify the computation, we used $\frac{\partial g(p)}{x} \approx g(p_x + 1, p_y, p_y) - g(p_x, p_y, p_y)$, $\frac{\partial g(p)}{y} \approx g(p_x, p_y + 1, p_z) - g(p_x, p_y, p_z)$ and $\frac{\partial g(p)}{z} \approx g(p_x, p_y, p_z + 1) - g(p_x, p_y, p_z)$ to approximate the spatial gradients. Then, we combined all loss terms to obtain the objective function as follows:

$$L = L_{wMAE} + \alpha L_{smooth} \tag{3}$$

where $\alpha$ is the weight for the smooth loss term. The model trained with Equation (3) as the objective function is referred to as the wMAE model. In training, $\alpha$ was set to 0.01.

Since a lung CT image typically contains multiple structures with distinct HU numbers, it is challenging to recover all structures simultaneously. Thus, we further extended wMAE model by applying a structure loss term to each of the clinical target volume (CTV) and five organs at risk (OARs), namely esophagus, heart, left lung, right lung, and cord. To be specific, the contours of both CTV and OARs were converted to bitmaps with 1 indicating the voxels within the ROIs and with 0 indicating the voxels outside the ROIs. Then, the bitmaps of *sCT* structures were generated by wrapping the bitmaps of *vCT* structures based on the generated DVF. Last, we used structural similarity index matrix (SSIM)[88] as the structure loss to compare the similarity between the bitmaps of *sCT* and *iCT* structures. SSIM was applied since it considers not only the similarity between the corresponding structures but also the illumination and contrast of the structures. On the contrary, the commonly used dice similarity coefficients (DSCs)[22,89,90] only considers the volume overlap of the structures. The SSIM is more appropriate for the lung CT images since the lung CT images often have multiple structures with a large range of the HU numbers that potentially leads to diverse illuminations and contrasts. The definition of SSIM is defined in Equation (4) as follows:

$$L_{SSIM}(s_{iCT}, s_{sCT}) = \frac{1}{K}\sum_{j=1}^{K} L_{SSIM}(s_{iCT}^{j}, s_{sCT}^{j})$$

$$L_{SSIM}(x, y) = \frac{(2\mu_x\mu_y + C_1)(2\sigma_{xy} + C_2)}{(\mu_x^2 + \mu_y^2 + C_1)(\sigma_x^2 + \sigma_y^2 + C_2)} \quad (4)$$

where $s_{iCT}$ represents the structures in *iCT* and $s_{sCT}$ represents the structures in *sCT*, respectively. *K* is the number of structures we considered in this study. $\mu_j$ and $\sigma_j$ represents the mean and standard deviation of the voxels in the structure *j*, respectively. $C_1$ and $C_2$ are two constants that ensure stability when the denominator becomes 0. A SSIM value of 1 indicates the best agreement and a value of 0 indicates the worst agreement. Finally, we combined all loss terms to obtain the objective function as follows:

$$L = L_{wMAE} + \alpha L_{smooth} + \beta L_{SSIM} \quad (5)$$

where $\alpha$ and $\beta$ are the weights for different loss. The model trained with Equation (5) as the objective function is referred as the M+S model. In training $\alpha$ was set to 0.01 and $\beta$ was set to 0.1.

Considering the limited but diverse lung CT images used in this study and the small dimension size of each CT image, the model tends to either easily overfit to the dataset or cannot fully capture the detailed voxel information. Thus, we introduced a training strategy -- random mask to address the issue. In the training stage, for each batch, we randomly masked out a cube of size $m$ ( $m \times m \times m$) from both the *iCT* and *vCT*. An illustration of the random mask strategy is shown in Figure 4.

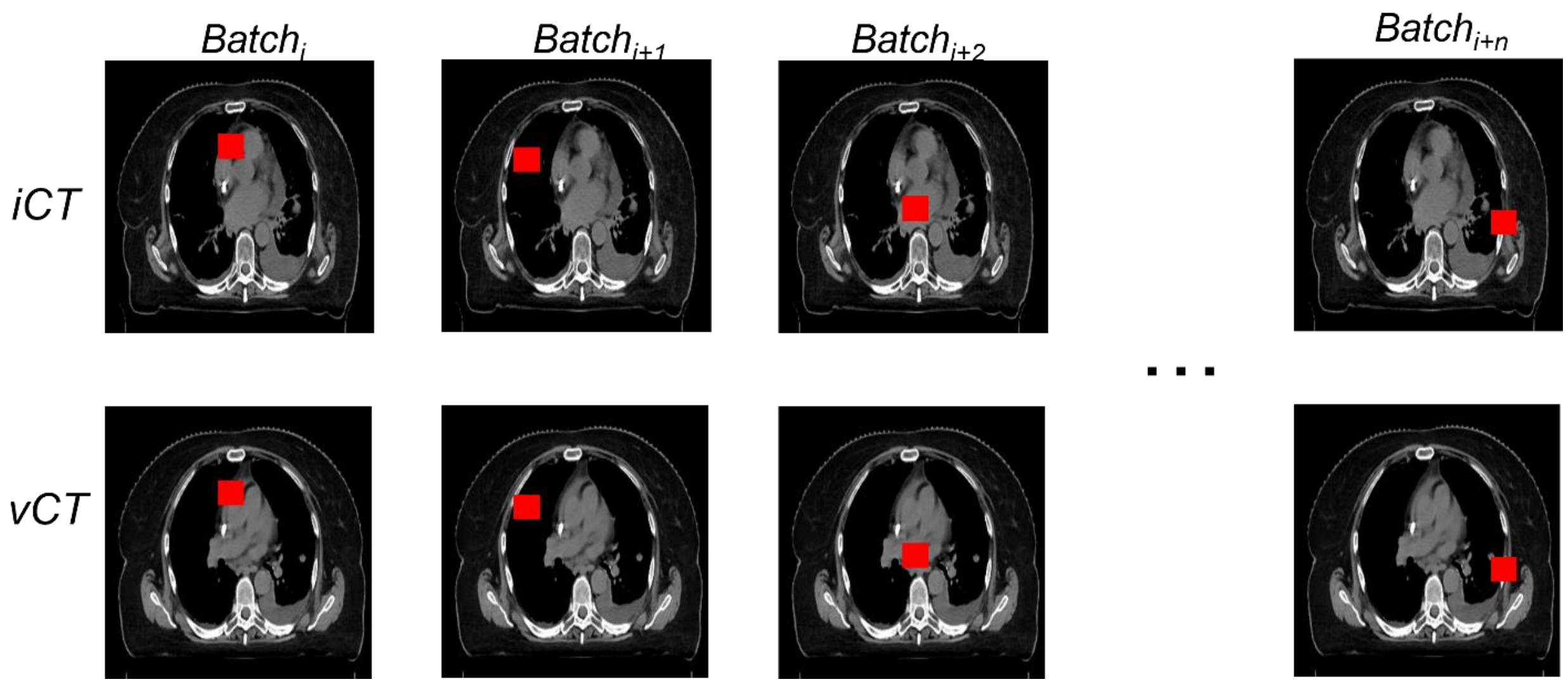

Figure 4. Illustration of the random mask strategy used in the model training. In each batch, the value of the voxels enclosed by the random mask will be set to 0.

Since the voxels in the masked cube were completely blocked out from the network in the given batch, the network would return higher losses for the masked cube, which would in turn make the model assign higher weights for the masked cube and yield images with better fine details in the next batch to reduce the loss. With the batches going on, the model would go through all the voxels and eventually result in good deformation for the entire image dataset. Moreover, the introduced random mask can also be considered as a way of data augmentation by inserting various noise (i.e., the masked cubic) for each batch, thus alleviating the risk of overfitting. Through an empirical study, we found that a random mask with a size of 5 balanced the accuracy and time cost the best, thus we used a size of 5 in the following experiments. Adam optimizer with an initial learning rate of 1e-4 was used for training and the hyperparameters associated with the Adam optimizer were $\beta1=0.9$ and $\beta2=0.999$. The models were implemented with the PyTorch

(https://pytorch.org/) deep learning library and the model were trained on four A100 GPUs with a batch size of 4. The proposed iterative training loop is summarized in Algorithm 1.

Algorithm 1: The training protocol of the proposed DIR method.

| | |
|---|---|
| **INPUT** | Pairs of initial CTs (*iCTs*) and verification CTs (*vCTs*), number of training epochs *N*, learning rate $\eta$, loss balance weights $\alpha$ and $\beta$, mask size *m*, the bitmaps of the CTV, and five OAR structures *s*. |
| **OUTPUT** | The synthetic CT (*sCT*), propagated contours, and the corresponding deformable vector fields (DVFs). |
| **FOR** | $n = 1$ to *N* **do** |
| | Randomly cropping a cubic of $m \times m \times m$ from both *iCT* and *vCT* |
| | Training the model with an objective function described in Equation (3) or (5) |
| | Adjusting the learning rate as $\eta = \frac{\eta}{10}$ if $N \bmod 50$ equals $0$. |
| **END** | |
| **RETURN** | *sCT,* propagated contours (bitmaps), and DVFs. |

In the validation stage, we would not apply the random mask to the given test pairs of lung CT images. The model would produce the DVFs given the test CT images pair and then generate the *sCT* as mentioned before. Moreover, the bitmaps of the CTV and any given OARs contours in *sCT* images were generated by warping the corresponding bitmaps of structure contours from *vCT* images based on the generated DVFs. The details of the validation steps are shown in Algorithm 2.

Algorithm 2: The validation protocol of the proposed DIR method.

| | |
|---|---|
| **INPUT** | Pairs of initial CTs (*iCTs*) and verification CTs (*vCTs*) in the testing dataset, bitmaps of ITV/PTV, five OARs structures *s* and the trained model ***M***. |
| **OUTPUT** | The synthetic CT (*sCT*), propagated contours (bitmaps), and the corresponding deformable vector fields (DVFs). |
| **FOR** | **ANY** *iCT* and *vCT**s*** in the testing dataset **do** |
| | Inputting them to the trained model *M* |
| **END** | |
| **RETURN** | *sCT*, propagated contours (bitmaps), and DVFs. |

## 2.4. Data Analysis

Both the trained wMAE and M+S models were validated in the testing dataset, which comprised of 10 independent patients and were applied with the same data pre-processing and data post-processing mentioned before. For image quality evaluation in the testing dataset, we directly measured the similarity using Per-voxel absolute CT-number-difference volume histogram (CDVH) and MAE as the evaluation metrics between the ground-truth CTs, i.e., *iCT*, and the synthetic CTs (*sCT*), which were derived by deforming the *vCT* with the derived DVFs.

Four conventional DIR approaches (fast symmetric force, diffeomorphic, log domain diffeomorphic, symmetric log domain diffeomorphic) were also compared. For the conventional DIR methods, we used the DIR algorithms included in the open source image registration library, Plastimatch[91], to register *vCT* to *iCT*. We used the same pre-processing procedure for the conventional DIR approaches for fair comparison.

For the dosimetric evaluation, we postprocessed both the *sCT* and *iCT* by inversing all the steps in the pre-processing stage, so that all *sCT*s and *iCT* had the same configurations as their corresponding *iCTo*s. Then, forward dose calculations of the original plan were done based on *sCT* and *iCT*. The resulting dose distributions were compared using the 3D Gamma analysis. Dose volume histograms (DVHs) were generated as well for these two dose distributions. We also compared the clinically relevant DVH indices for the selected structures. We considered $D_{95}$ and $D_5$ (the minimum dose covering the highest irradiated 95% and 5% of the structure's volume, respectively) for CTV, $V_5$ (the minimum volume percentage receiving at least 5Gy [RBE]) for total lung, $D_{mean}$ (mean dose) for heart, $D_{max}$ (max dose) for cord, and $D_{mean}$ for esophagus. The clinically relevant DVH indices were also statistically analyzed using the paired Student's *T*-test. A $P$-value $\leq 0.05$ was considered to be statistically significant.

## 3. Results

Section 3.1 and Section 3.2 report the evaluation regarding the image quality of the *sCT*. Section 3.3 and Section 3.4 show the dosimetric evaluation on the *sCT*.

### 3.1. Evaluation of the *sCT* quality

Table 1 displays the comparison of the HU number MAE and time cost of the proposed approach and four conventional DIR approaches (fast symmetric force, diffeomorphic, log domain diffeomorphic, symmetric log domain diffeomorphic). From Table 1, we observed that the proposed methods achieved better quality *sCT* as indicated by much smaller MAE compared to the conventional methods with a time cost of fewer than 300 milliseconds whereas all conventional DIR approaches suffered from worse quality *sCT* (as indicated by larger MAE) and all with a much longer (at least 1000 times larger than our methods) computation time.

Furthermore, we derived the per-voxel CT-number absolute difference volume histogram (CDVH) with the absolute CT number differences (in HU) as the horizontal axis and the normalized volume (in %) as the vertical axis (Figure 4) to show the per-voxel absolute CT-number difference between *sCTs* and *iCT* statistically. As shown in the figure, a majority of the voxels had a small per-voxel absolute CT number difference (close to 0) between the *iCT* and *sCT*. Statistically, only 5% of the voxels have a per-voxel absolute CT number difference larger than 46.7538 HU for the model trained with the weighted MAE only and 54.6117 HU for the model trained with both the weighted MAE and SSIM, respectively.

Table 1. Comparison of the MAE and time cost of the proposed approaches and four conventional DIR approaches. In each cell, we reported the mean and standard deviation value of 10 test patients.

| | MAE (HU) | Time cost (Seconds) |
|---|---|---|
| wMAE | 13.15±3.8 | $(263.7\pm 163)\times10^{-3}$ |
| M+S | 17.52±5.8 | $(265.8\pm 190)\times10^{-3}$ |
| FSF | 56.4±18.1 | 280.3±129.8 |
| DM | 100.4±25.5 | 283.5±125.2 |
| LD | 249.0±59.4 | 290.9±101.5 |
| SLD | 400.96±69.4 | 304.3±97.2 |

*abbreviations*: FSF for fast symmetric force, DM for diffeomorphic, LD for log domain diffeomorphic, SLD for symmetric log domain diffeomorphic

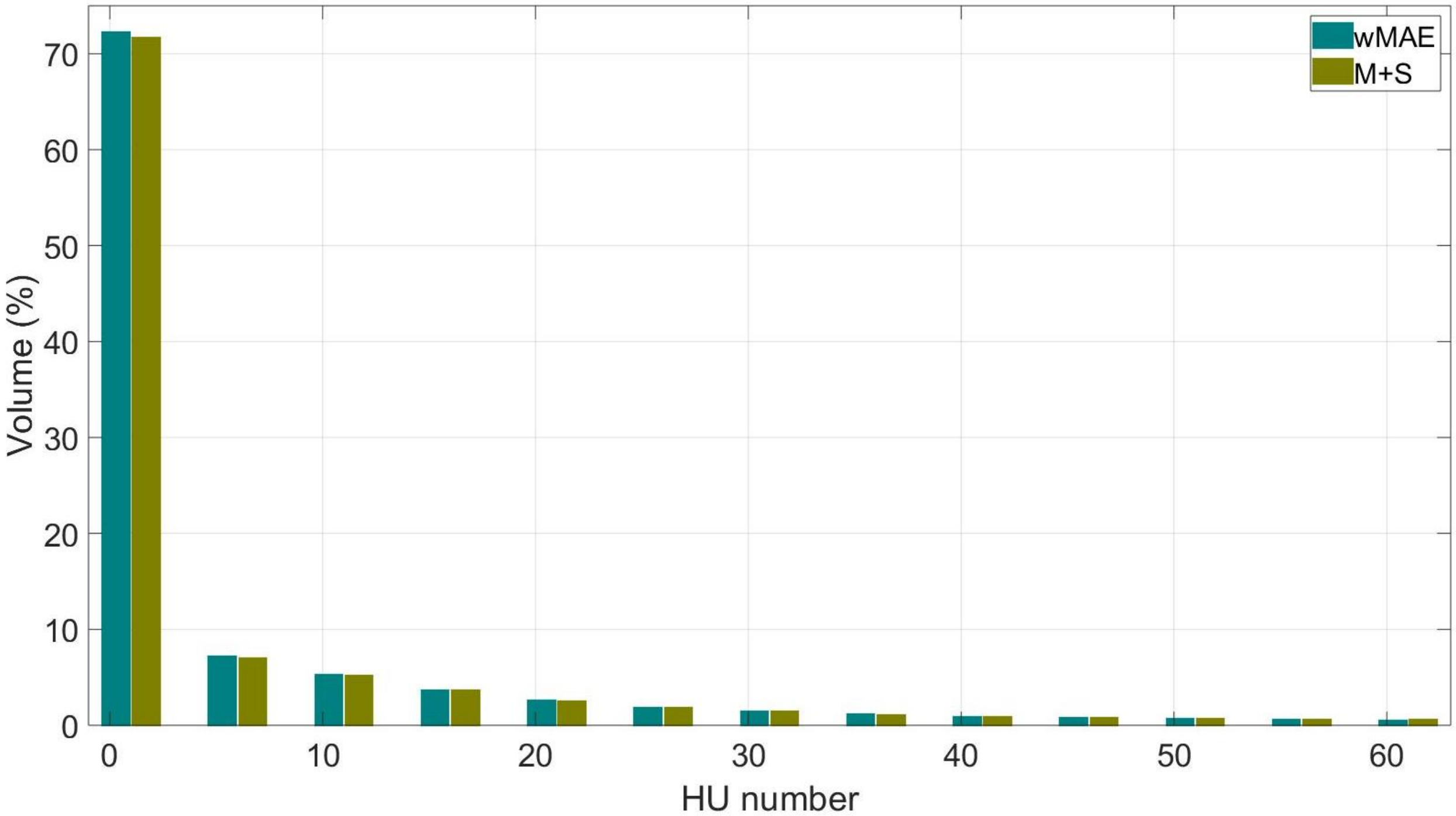

Figure 4. The absolute per-voxel CT number difference volume histogram for a typical patient. The x-axis represents the HU number absolute difference between the *sCT* and *iCT* and the y-axis represents the percentage of the volume. For both models, more than 70% of the volume are exactly reconstructed.

Figure 5 compares a typical CT slice between the *iCT* (*a*) and the corresponding *sCT* generated by the wMAE model (*b*) and the M+S model (*c*), respectively. The differences between the *iCT* slice and *sCT* slices are shown in Fig. 5(*d*) and (*e*), where the brighter the color is, the greater the difference is. Overall, both *sCT* matched the *iCT* well and no obvious artifacts were identified. There were, however, some discrepancies between the *sCT* and the *iCT* in some soft tissue regions. When comparing the *sCT* (wMAE) and the *sCT* (M+S), the *sCT* (wMAE) achieved a lower average MAE and had much smoother edges (skin), although its image quality appeared slightly worse than the *sCT* (M+S) since the sternum was blurred (blue rectangle in Figure 5) and the high Z material in muscle was barely recovered (green rectangle in Figure 5).

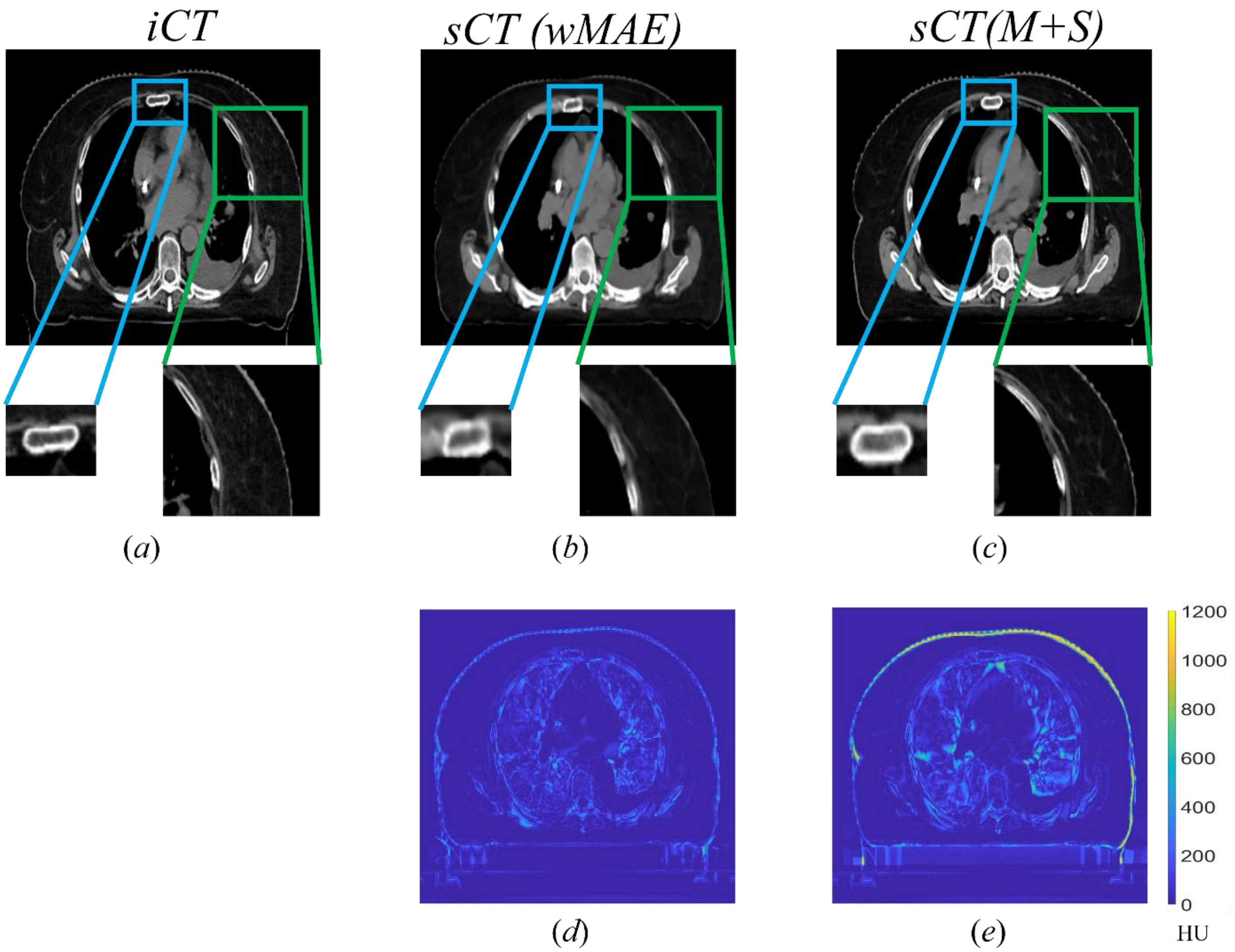

Figure 5. Comparison of a typical slice between the *iCT* slice (*a*) and its corresponding *sCT* generated by the model trained with wMAE only (wMAE) (*b*) and the model trained with both wMAE and SSIM (M+S) (*c*), respectively. The CT HU number display window level and width were -125HU and 1300HU respectively. The edge of the rectum was more blurred in *sCT* generated by the wMAE model. The high Z material in *sCT* generated by the M+S model were partially recovered, whereas the wMAE model did not. Fig. 5(*d*) and (*e*) showed the absolute

difference of the slice between the *iCT* and *sCT* generated by the wMAE and (M+S) models, respectively.

### 3.2. Evaluation of the propagated contours

Similar to the procedure in section 3.1, we generated a new set of contours by propagating the contours from the *vCT* to *iCT* based on the derived DVFs from both trained models. Then we measured the similarity between the propagated contours and the initial contours using SSIM. The detailed results are shown in Table 2.

Table 2. Comparison of SSIMs of the selected structures generated by the models trained with wMAE and M+S, respectively. The higher the SSIM value is, the higher the agreement of the selected structures between the *sCT* and *iCT* is.

| | wMAE | M+S |
|---|---|---|
| CTV | 0.989±0.013 | 0.993±0.009 |
| Right lung | 0.973±0.011 | 0.975± 0.008 |
| Left lung | 0.976±0.008 | 0.977±0.007 |
| Esophagus | 0.997±0.001 | 0.997±0.001 |
| Heart | 0.987±0.005 | 0.989±0.005 |
| Cord | 0.998±0.001 | 0.998±0.001 |
| average | 0.987±0.006 | 0.988±0.004 |

From the table, it is obvious that the proposed methods can successfully generate contours with excellent agreement for the selected structures (CTV, right lung, left lung, esophagus, heart, and cord) with the ground-truth contours after propagation (the average SSIM scores of 0.987$\pm$0.006 and 0.988$\pm$0.004 for the two proposed models, respectively). The model trained with M+S achieved higher SSIM scores for all structures than those of the model trained with wMAE. We also calculated the DSCs, which only consider the overlap between the ground-truth and propagated structures, for the selected structures. We found that both left and right lung suffered from low DSC scores (approximately 0.75) due to inter-fractional anatomy changes and irregular respiratory patterns, whereas the SSIM score was not greatly affected by the anatomy changes.

### 3.4. Comparison of the dose volume histograms (DVHs) and the clinically relevant DVH indices

We performed the forward dose calculation of the original plan on the *iCT* and *sCT*. We then generated dose volume histograms (DVHs) based on the two dose distributions for every testing patient. Figure 6 shows the comparison of the DVHs generated from the dose distributions calculated on the *iCT* and its corresponding *sCT* for a typical photon patient, where the red curve represents the CTV, the blue curve represents the total lung, the purple curve represents the heart, the green curve represents the cord and the magenta curve represents the esophagus. The solid, dashed, and dotted represent the DVHs generated from the dose distributions calculated based on *iCT*, *sCT*(wMAE) and *sCT*(M+S), respectively. For better visualization, zoom-in detailed subfigures and DVH indices differences were also provided. Visually, the DVH curves on *iCT* and *sCTs* completely overlapped with each other with negligible differences only visible in the zoomed-in regions.

Figure 7 shows the comparison of the boxplots of the clinically relevant DVH indices of the ten testing patients from the dose distributions calculated on *iCT* and the corresponding *sCTs* derived from the two models proposed in this study. *P*-values are shown on the top of the boxplots. From Figure 7, it is clear that the clinically relevant DVH indices derived from the dose distributions calculated on the *sCT*s were very similar to the ones from the dose distribution calculated on *iCT* for all selected structures.

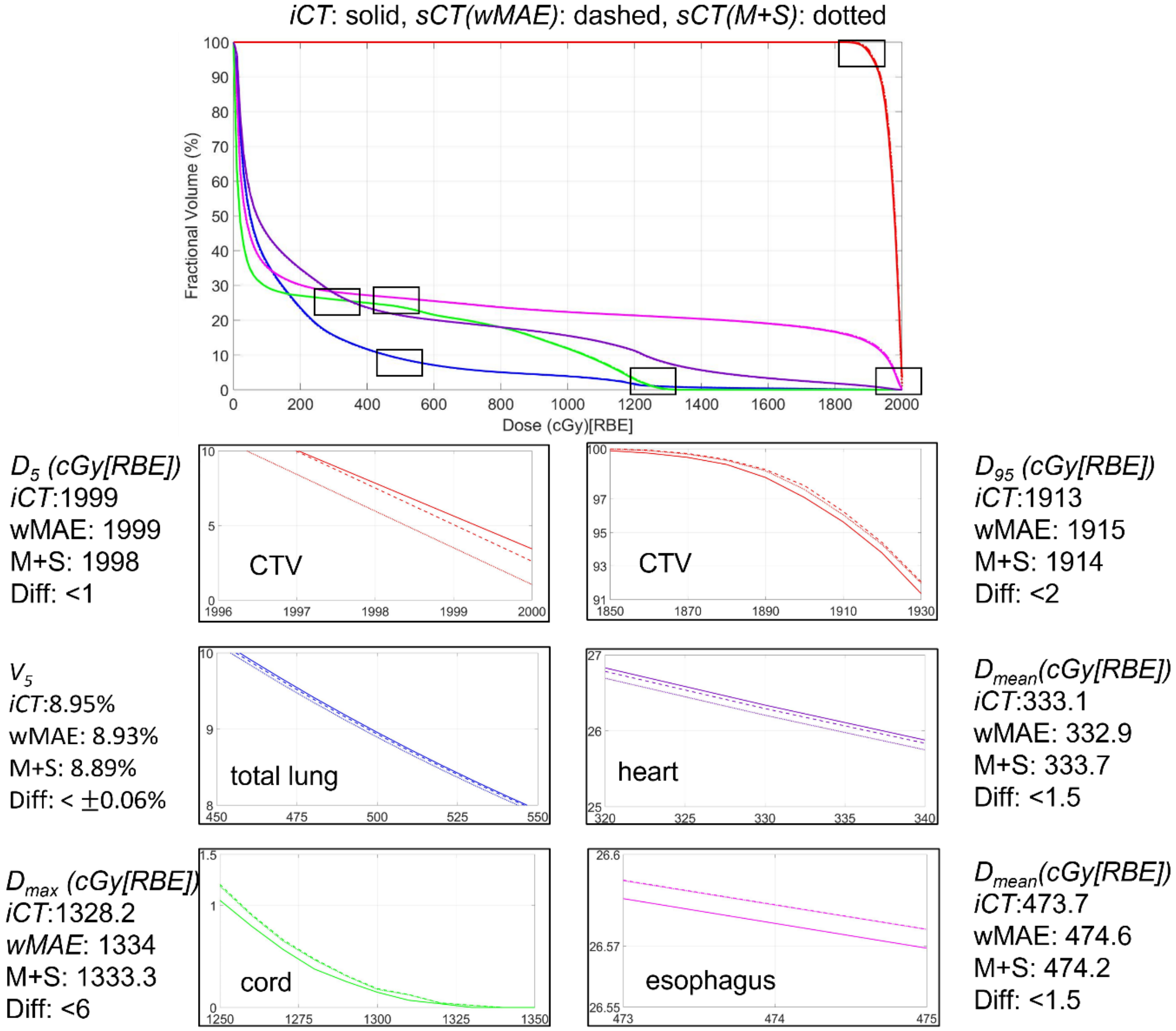

Figure 6. Comparison of dose volume histograms (DVHs) of one typical patient derived from the dose distributions calculated on *iCT* and the corresponding *sCTs* derived from the two models proposed in this study. In each figure, the solid, dashed, and dotted lines represent the DVHs generated from the dose distributions calculated based on *iCT*, *sCT*(wMAE) and *sCT*(M+S), respectively.

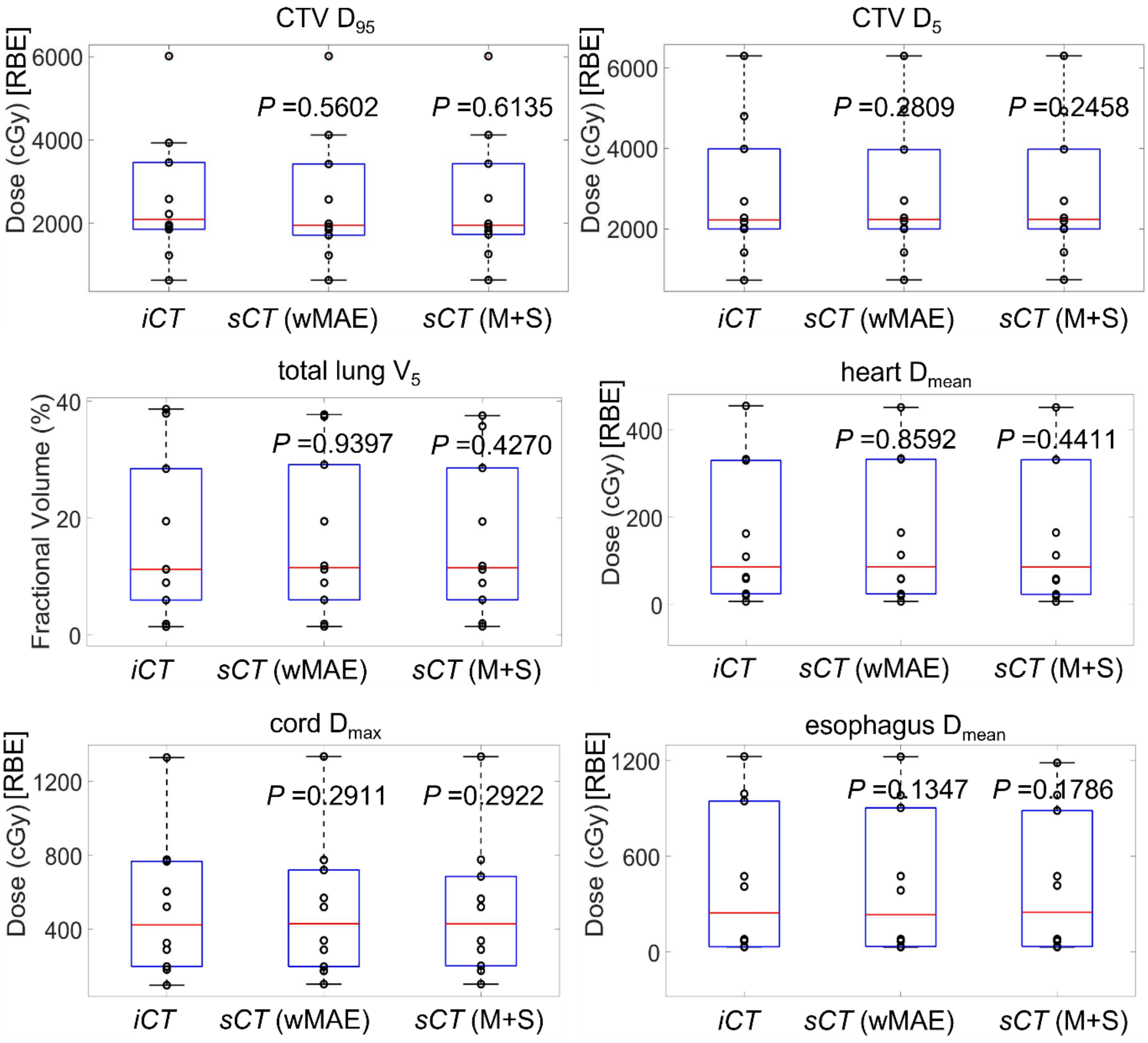

Figure 7. Comparison of the boxplots of the clinically relevant DVH indices of the ten testing patients from the dose distributions calculated on *iCT* and the corresponding *sCTs* derived from the two models (wMAE and M+S) proposed in this study. *P*-valves derived from the statistic tests between the DVH indices calculated based on the dose distributions on *iCT* and *sCT* (wMAE), and between the DVH indices calculated based on the dose distributions on *iCT* and *sCT* (M+S) are shown on the top of the corresponding boxplots.

### 3.5 Comparison of the dose distributions using 3D Gamma analysis

We also compared the dose distributions calculated on *iCT* and the corresponding *sCTs* derived from the two models proposed in this study using 3D Gamma analysis with a threshold of 3%/3mm/10% and 2%/2mm/10%, respectively (Table 3). The average 3D gamma passing rate for a threshold of 3%/3mm/10% was above 98% and above 96% for the wMAE model and the M+S model, respectively. For a threshold of 2%/2mm/10%, the average 3D Gamma passing rate was above 97%.and above 94% for the wMAE model and the M+S model, respectively. If the 10$^{th}$ testing patient was excluded, the average 3D Gamma passing rate for the remaining 9 testing cases was above 99% for a threshold of 3%/3mm/10% and above 98% for a threshold of 2%/2mm/10% for the wMAE model. Moreover, the average 3D Gamma passing rate for patients treated with photon therapy was approximately 3% higher than that of patients treated with proton therapy. We also noticed that the wMAE model obtained a higher 3D Gamma passing rate than that of the M+S model.

Table 3. The 3D Gamma passing rates between the dose distributions calculated on *iCT* and the corresponding *sCTs* of the 10 testing patients with a threshold of 3%/3mm/10% and 2%/2mm/10% for both the wMAE model and the M+S model, respectively. * indicates that the patient was treated with proton therapy.

| **PATIENT** | **3%/3mm/10%** (wMAE) / (M+S) | **2%/2mm/10%** (wMAE) / (M+S) |
|---|---|---|
| **#1** | 0.999 / 0.998 | 0.997 / 0.996 |
| **#2*** | 0.996 / 0.965 | 0.98 / 0.926 |
| **#3** | 1.0 / 0.991 | 0.999 / 0.977 |
| **#4** | 1.0 / 0.999 | 0.999 / 0.999 |
| **#5** | 0.995 / 0.943 | 0.978 / 0.913 |
| **#6*** | 0.966 / 0.976 | 0.921 / 0.934 |
| **#7** | 1.0 / 0.979 | 1.0 / 0.970 |
| **#8** | 1.0 / 0.979 | 1.0 / 0.970 |
| **#9** | 0.999 / 0.924 | 0.991 / 0.811 |
| **#10*** | 0.921 / 0.924 | 0.899 / 0.902 |
| **Average** | 0.986±0.026 / 0.963±0.029 | 0.977±0.036 / 0.945±0.062 |
| **Average(photon)** | 0.995±0.012 / 0.974±0.029 | 0.987±0.029 / 0.955±0.067 |
| **Average(proton)** | 0.971±0.043 / 0.944±0.020 | 0.953±0.047 / 0.907±0.007 |

## 3.6 Ablation Study

Figure 8 showed the *iCT* and corresponding *sCT*s generated by different models trained with different loss terms. Figure 8 (*a*) and (*b*) show the *sCTs* generated by the model trained with the SSIM or the MSE loss term only, respectively. Figure 8 (*b*) was the model setting adpoted by the original Voxelmorph model. Figure 8 (*c*) presented the result generated by the model trained with both the MSE and SSIM loss terms, Figure 8 (*d*) was for the model trained with the wMAE loss term without applying ramdon mask strategy and Figure 8 (*e*) was the iCT. Figure 8 (*f*)-(*i*) were the absolute difference between the *sCT*s and *iCT*. Compared with the results showed in Figure 5, it was clearly seen that no models shown in Figure 8 generated the *sCT*s with better details, for example, neither of the models can generate sCT with corrected rectum nor cord. We further quantitatively evaluated the quality of the *sCTs* by different models. The *sCT*s shown in Figure 8 (*a*), (*b*), (*c*) and (*d*) achieved a MAE of 52.73 HU, 26.12HU, 19.42HU and 22.64HU, respectively (Table 4), while the wMAE and M+S models proposed in this work achieved a MAE of 13.15HU and 17.52HU, respectively (Table 1).

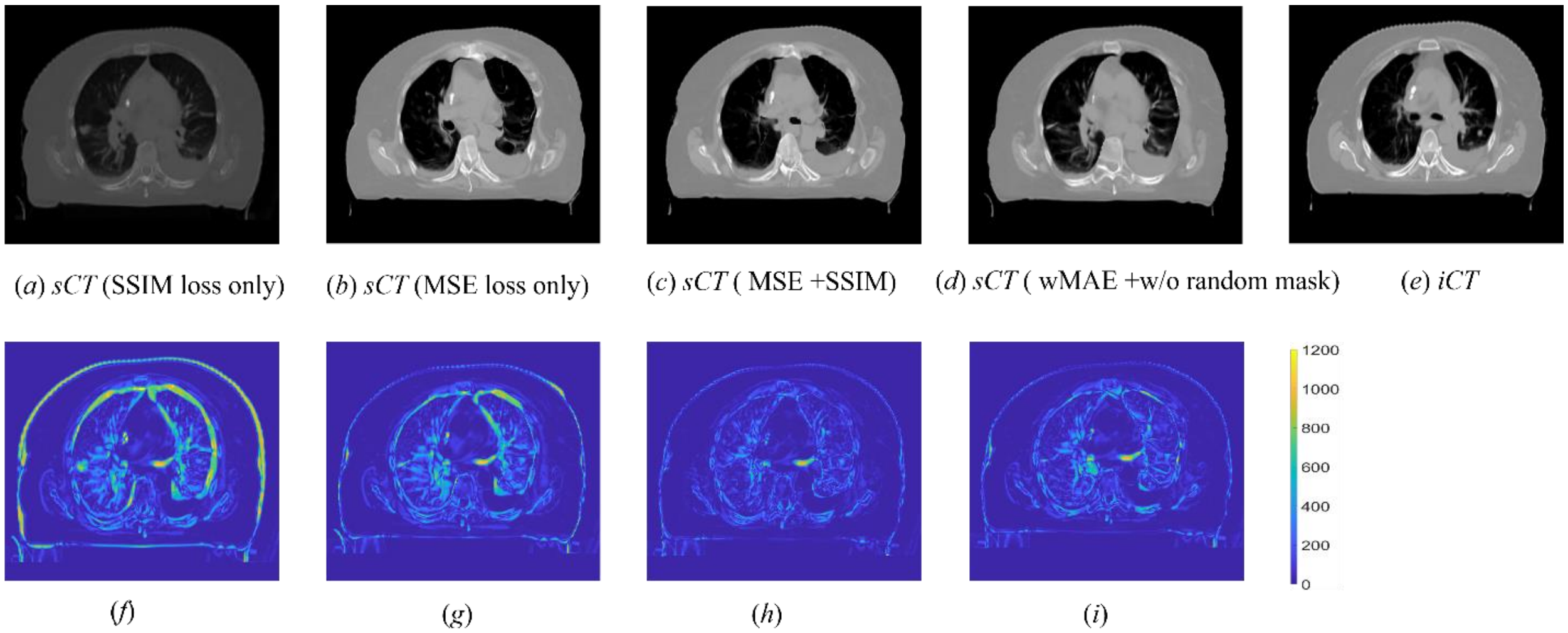

Figure 8. Sample slices of the *sCTs* generated by the models trained with different loss terms and the corresponding *iCT* (*e*). Figure (*a*) and (*b*) are the sample slices of the *sCTs* generated by the models trained with the SSIM or MSE loss term only, respectively. Figure (*c*) shows the sample

slice of the *sCT* generated by the model trained with both the MSE and SSIM loss terms. Figure (*d*) shows the sample slice of the *sCT* generated by the model trained with the wMAE loss term without random mask strategy. The CT HU number display window position and width were -120 HU and 1300 HU, respectively. Figure (*f*)-(*i*) show the corresponding absolute differences between the *sCT*s and *iCT*.

Table 4. The MAE of the sCT generated by models trained with different loss terms.

| | MAE(HU) |
|---|---|
| SSIM loss only | 52.73 |
| MSE loss only | 26.12 |
| MSE +SSIM | 19.42 |
| wMAE w/o random mask | 22.64 |

**4. Dicussion**

In this study, we have developed a VoxelMorph-based deep neural network for fast and accurate DIR in the radiotherapy of lung cancer. We tried two configurations (thus two models) in the proposed methods and performed a comprehensive validation of the proposed models on the CT images of lung cancer patients. Although the methods based on CT images were focused on lung cancer, the methods could be generalized to all disease sites.

To alleviate the potential overfitting caused by limited data and low resolution, we introduced an random mask training strategy, and included additional loss terms in the objective function (i.e, the weighted MAE and SSIM terms), to improve the quality of the *sCT*. Through an empirical study, we found that a random mask with a size of $5 \times 5 \times 5$ yielded the optimal performance in our study. A random mask with a very small size cannot mitigate the blur in the *sCT* images well enough, whereas a random mask with a very large size, though it may help to mitigate the blur, introduces uncertainty to  the model training and eventually slows down or even collapses the training of the neural network. As for the loss terms, we used the weighted MAE to guarantee the voxel-to-voxel similarity, in which the per-voxel loss weight is propotional to the HU numbers, thus helping to reduce the high-frequency artifacts (e.g., the artifacts in bone structure). Comparing with the MSE loss term, which is used by the original Voxelmorph model, the use of the wMAE loss term greatly improves the quality of the *sCT*. Another additional loss term used in this study is the SSIM, which is a loss that enforces the similarity among structures. However, unlike dice similarity coefficients (DSCs)[22,89,90], which has been extensively used as the structure similarity evaluation metric, SSIM considers not only the similarity among structures, but also the illumination and contrast among the images. Thus it is a better choice for the lung CT images since the lung CT images often have multiple structures with large variation of the HU numbers. This potentially leads to diverse illuminations and contrasts.

We trained the proposed neural network with two configurations (thus two models), one with the weighted MAE loss term only (wMAE model) and the other with both the weighted MAE loss term and SSIM loss term (M+S model). The results related to *sCT* quality during the validation showed that the M+S model yielded slightly better *sCT* image quality than that of the wMAE model. However, the dosimetric evaluation by comparing the clinically relevant DVH indices and performing 3D Gamma analysis between the dose distributions calculated on *iCT* and the corresponding *sCTs* derived from the two models proposed in this study presented the opposite results – the wMAE model had a slightly better agreement of the DVH indices with ground-truth and a higher 3D Gamma passing rate. This indicates that the evaluation of the *sCT* quality needs to be conducted thoroughly and cannot rely on one criterion alone. Additionally, it further suggests that high *sCT* image quality, although clinically relevant in radiation therapy, does not necessarily lead to favorable results in dose calculation. How to further improve the performance of the proposed deep neural network in dose calculation will be an interesting and challenging research direction.

The 3D Gamma passing rates reported in Table 3 are very promising and exceeds the clinical requirements suggested by American Association of Physicists in Medicine (AAPM) task group (TG) 218[92], suggesting that the generated *sCT*s can be reliably used in clinical applications such as ART. However, it requires further improvements. There are multiple factors for these non-ideal results. One factor is inherited in the different physics characteristics of proton and photon therapy, where proton therapy could be more sensitive to the same variations in the HU numbers compared to photon therapy as shown in Table 3. Another possible contributing factor can be the challenging CT imaging dataset used in this study: all the CT images pairs (*iCT* and *vCT*) used in this study are the CT images taken at least several weeks apart, during which the inter-fractional anatomy changes can be large, irregular and unpredictable. In addition, the tumor may grow or shrink and patients' weight may change during the time window between the *iCT* and *vCT*, thus introducing additional unpredictable ambiguities (new information or loss of the old information compared to the information contained in *iCT*) for DIR. Figure 9 shows the comparison of one CT slice in the middle of the tumor among the *iCT*, *vCT* and *sCT*s of a case with a relatively poor performance from our proposed methods (the 10$^{th}$ testing patient in Table 3), where the circle highlights the CTV in each subfigure. It is obvious that the generated *sCT*s have a CTV with larger high density regions compared to the groud-truth CT (i.e., *iCT*), possibly due to the fact that this patient has a very aggressive tumor phenotype (the tumor grows a lot from *iCT* to *vCT*). This might lead to worse agreements of DVH indices and a lower 3D Gamma passing rate (92.11% of 3mm/3%/10% and 89.92% of 2mm/2%/10% for the wMAE model and 92.30% of 3mm/3%/10% and 90.21% of 2mm/2%/10% for the M+S model, respectively). Moreover, for more challenging disease sites which have complexity shapes, such as ovarian cancer[94] and for registration between different modalities[95], the proposed DIR approach may see its limitation. Further investigation are needed to address these issues.

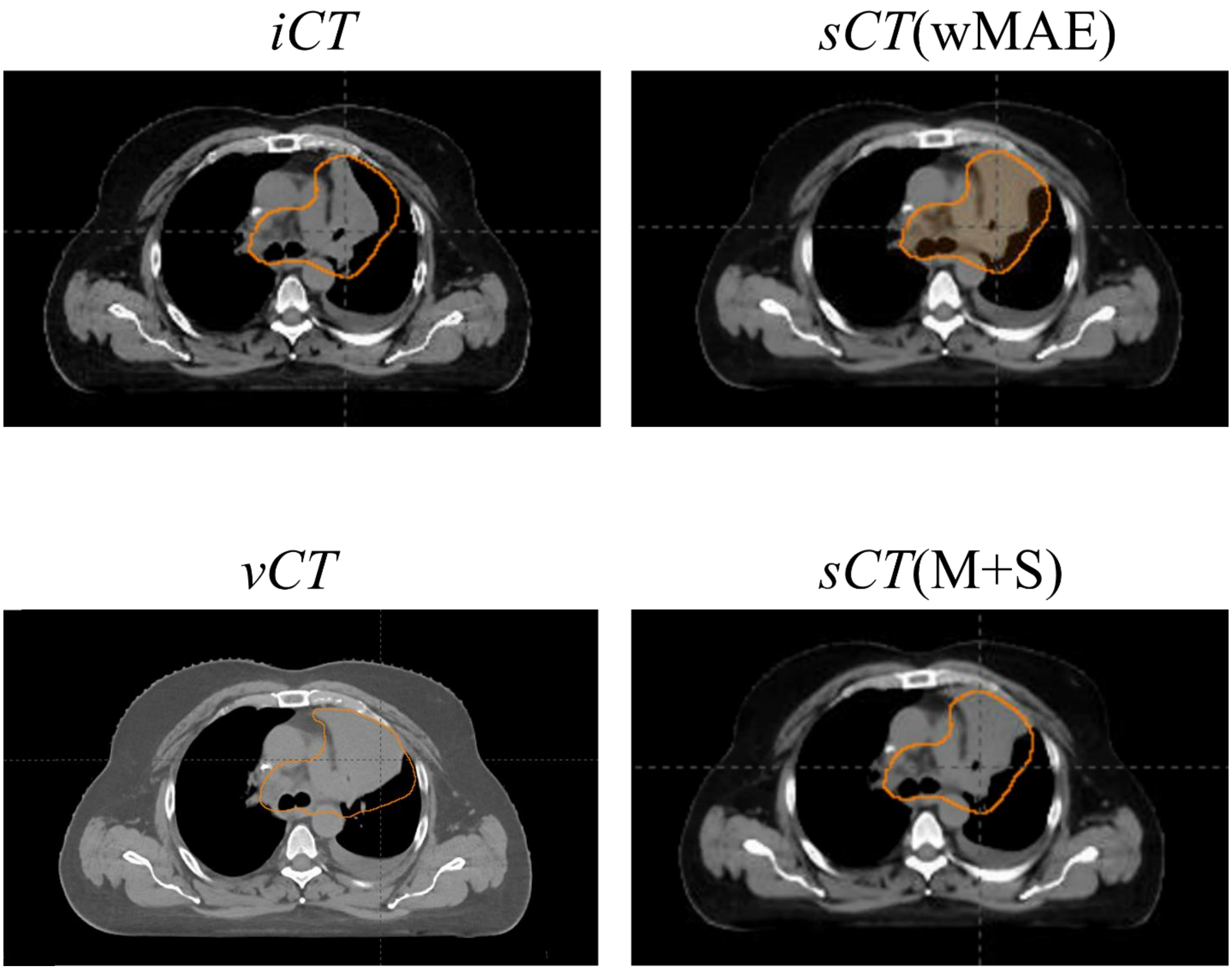

Figure 9. Comparison of one CT slice in the middle of CTV among the *iCT*, *vCT* and *sCT*s of a case with a relatively poor performance from our proposed methods (the 10$^{th}$ testing patient in Table 3), where the circle highlights the CTV in each subfigure.

## 5. Conclusion

A deep neural network-based DIR approach was proposed and shown to be accurate and efficient to register the initial CTs and verification CTs for lung cancer.